\documentclass[11pt,a4paper]{article}
\usepackage[hyperref]{emnlp-ijcnlp-2019}
\usepackage{times}
\usepackage{latexsym}
\usepackage{xcolor}
\usepackage{url}
\usepackage{todonotes}
\usepackage{graphicx,subcaption}
\usepackage{xspace}
\aclfinalcopy 

\newcommand\system{\textit{ALTER}\xspace}

\newcommand{\task}{\textit{GMETCA}\xspace}
\newcommand{\tool}{\system}

\title{ALTER: Auxiliary Text Rewriting Tool for Natural Language Generation}

\author{Qiongkai Xu \ \ \ \ Chenchen Xu\\
  The Australian National University \\
  Data61 CSIRO \\
  {\tt Qiongkai.Xu@anu.edu.au} \\
  {\tt Chenchen.Xu@anu.edu.au} \\\And
  Lizhen Qu \\
  Laboratory for Dialogue Research\\
  Monash University \\
  {\tt Lizhen.Qu@monash.edu} \\}

\date{}

\begin{document}
\maketitle
\begin{abstract}
  In this paper, we describe \system{}, an auxiliary text rewriting tool that facilitates the rewriting process for natural language generation tasks, such as paraphrasing, text simplification, fairness-aware text rewriting, and text style transfer. Our tool is characterized by two features, i) recording of word-level revision histories and ii) flexible auxiliary edit support and feedback to annotators. The text rewriting assist and traceable rewriting history are potentially beneficial to the future research of natural language generation.
\end{abstract}

\section{Introduction}
Generative modeling of editing text with respect to control attributes, coined \task, has seen increasing progress over the past few years. Such a generative task is referred to as \textit{style transfer}, when the control attributes indicate a change of writing styles~\cite{NAACL19_eval_style,fu2018style}. This generative task subsumes also gender obfuscation~\cite{reddy2016obfuscating}, authorship obfuscation~\cite{shetty18_a4nt}, and text simplification~\cite{xu2015problems}, when the control attributes indicate protection of gender information, protection of authorship, and simplifying the content and structure of the text, respectively. 

The research on \task are impeded by the lack of standard evaluation practices~\cite{NAACL19_eval_style,Tikhonov18_what_is_wrong}. Different evaluation methods make system comparison across publications difficult. In light of this, \newcite{NAACL19_eval_style,fu2018style} proposed both human evaluation and automated methods to judge style transfer models on three aspects: a) style transfer intensity; b) content preservation; c) naturalness. However, it is still difficult to reach an agreement on how to measure to what extent a generated text satisfy all three criterion. Moreover, the lack of human generated gold references hinders the progress of related research, as they i) automate error analysis as in~\cite{NAACL18_retrive_style_transfer}; ii) avoid repeated efforts in user studies to check if system outputs reproduce human-like editing. Therefore, it is beneficial to collect gold references, human edited text, as test corpora for those emerging tasks. 

The collection of gold references can be conducted on a crowd-sourcing platform, such as Amazon Mechanical Turk\footnote{\url{https://www.mturk.com/}}, or through existing writing tools~\cite{ACL19_demo_story_generation}. However, the existing crowd-sourcing platforms and annotation tools do not have the flexibility to add task-specific classifiers and language models, which are widely used for evaluating \task models~\cite{NAACL19_eval_style}.
As pointed out by~\newcite{dow2011shepherding}, it is important to incorporate task-specific feedback to achieve the improvement of user engagement and quality of results. Feedback is particularly important for \task according to our user study (details in Section \ref{sec:human_vs_machine}), because annotators fail to capture the weak associations between certain textual patterns and attribute values. For example, for gender obfuscation on \textit{`The dessert is yummy !'}, people can easily overlook the implicit indicator \textit{`yummy'} of female authors.

\begin{table}[h]
\centering
\small
\begin{subtable}[]{\linewidth}
\begin{tabular}{p{0.2cm} l}
Ori: & My husband and I enjoy LA Hilton Hotel. \\ 
P$_1$: & Family enjoy LA Hilton Hotel. (Rs)\\
P$_2$: & Family enjoy Hilton Hotel in LA. (Ro)\\
P$_3$: & All family members enjoy Hilton Hotel in LA. (I)\\
P$_4$: & All family members love Hilton Hotel in LA. (Rv)\\
\end{tabular}
\caption{Revision history 1 (RH1)}
\end{subtable}

\begin{subtable}[]{\linewidth}
\begin{tabular}{p{0.2cm} l} \\
Ori: & My husband and I enjoy LA Hilton Hotel. \\ 
P$_1$: & My husband and I love LA Hilton Hotel. (Rv)\\
P$_2$: & My husband and I love Hilton Hotel. (D)\\
P$_3$: & My husband and I love Hilton Hotel in Los Angeles. (I)\\
P$_4$: & My husband and I love Hilton Hotel in LA. (Ro)\\
P$_5$: & Family love Hilton Hotel in LA. (Rs)\\
P$_6$: & All family members love Hilton Hotel in LA. (I)\\
\end{tabular}
\caption{Revision history 2 (RH2)}
\end{subtable}

\caption{Two revision histories, RH1 and RH2, from `My husband and I enjoy LA Hilton Hotel.' to `All family members love Hilton Hotel in LA.'. Although the overall transformations of RH1 and RH2 are similar, they follow different revision histories.}
\label{tab:edit_history}
\vspace{-2mm}
\end{table}



To tackle the aforementioned challenges, we design \tool, an \underline{\textbf A}uxi\underline{\textbf L}iary \underline{\textbf {TE}}xt \underline{\textbf R}ewriting tool, to collect gold references for \task. Our tool contains multiple models to provide feedback on rewriting quality and also allows easy incorporation of more task-specific evaluation models. In addition, our tool has a module to record word-level revision histories with edit operations. The revisions are decomposed into a sequence of word-level edit operations, such as insertions (I), deletions (D), and replacements (R), as illustrated in Table~\ref{tab:edit_history}. The benefits of revision histories are three-fold. Firstly, revision histories can provide supervision signals for the generative models, which consider rewriting as applying a sequence of edit operations on text~\cite{NAACL18_retrive_style_transfer,TACL18_edit_prototype}. Secondly, revision histories can potentially provide deep insights regarding cognitive process and human edit behaviours in varying demographic groups.
For example, in Table~\ref{tab:edit_history}, human writers could prefer replacing the subject (Rs) and the object (Ro) as RH1 than replacing the verb (Rv) as RH2. Statistics on revision histories could provide supporting evidence about related assumptions.
Thirdly, there are often multiple gold references for the same text. It is more accessible using revision histories to acquire multiple references than rewriting every reference from scratch. As shown in Table~\ref{tab:edit_history}, P3, P4 in RH1 and P1, P3, P4 and P6 in RH2 are all valid revisions of the original sentence.

To sum up, our contributions are:
\begin{itemize}
    \item We implemented a tool \tool, which is capable of providing instant task-specific feedback on rewriting quality for \task.
    \item \tool records revision histories with edit operations, which are useful for comparing and analyzing human edit behaviours.
\end{itemize}
The code of \tool is publicly available under MIT license at \url{https://github.com/xuqiongkai/ALTER}. A screencast video demo of our system is provided at \href{https://drive.google.com/file/d/1GS-kbhhJKMxM3pxBZJdn7sVxP9rAfe4w/view?usp=sharing}{Google drive}.

\section{Related Work}
Our work is related to the research on edit history of text and assistant text rewriting.

Document-level edit records were used as data to analyze the evolution of knowledge base~\cite{ACL11_demo_wiki_revision,medelyan2009mining} and retrieve sentence paraphrases~\cite{max2010mining}. In contrast, our work focuses on word-level edit operations with order. We believe such paradigm introduces more linguistic features, that will benefit both linguistic and social behavior research.
Recently, there has been a series of work on conducting edit operations on text to advance automatic natural language generation~\cite{TACL18_edit_prototype,NAACL18_retrive_style_transfer}. We believe the real-world human rewriting history collected by our system will strengthen these works.

A writing assistant has been proposed to facilitate users, organizing and revising their document.
\newcite{NAACL16_demo_argrewrite} proposed to detect the writers' purpose in the revised sentences.
\newcite{ACL19_demo_story_generation} developed a collaborative human-machine story-writing tool that assists writers with story-line planning and story-detail writing.
The assistant and feedback generally improved the user engagement and the quality of generated text in those works.

\begin{small}
\begin{figure*}[t]
    \centering
	\includegraphics[width=.9\linewidth]{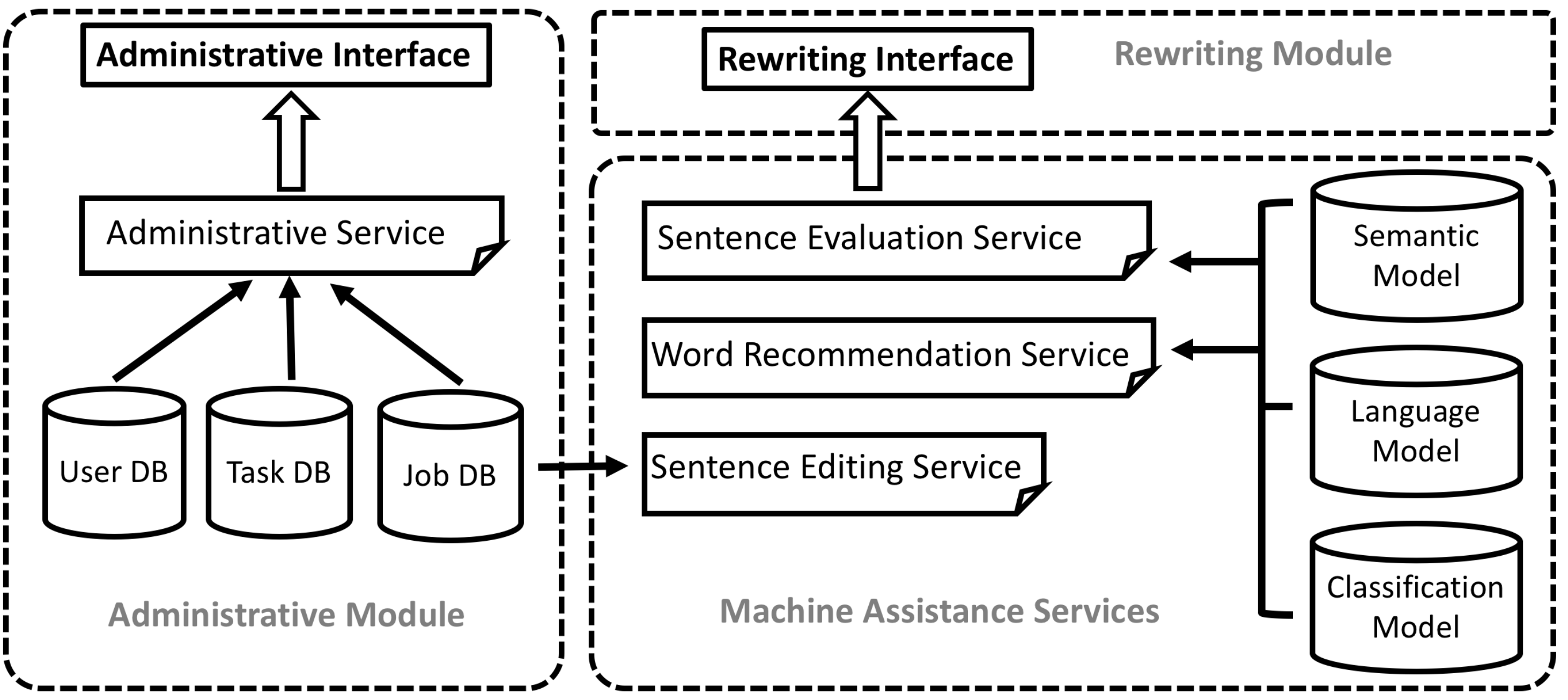}
	\vspace{-2mm}
	\caption{System architecture of the Auxiliary Text Rewriting Tool (ALTER).}
	\label{fig:system_archietecure}
	\vspace{-2mm}
\end{figure*}
\end{small}

\section{ALTER}
In this section, we describe the design of \system, an auxiliary text rewriting tool that is able to i) provide instant task-specific feedback to encourage user engagement, and ii) trace the word-level revision histories. We demonstrate an example of adapting our system on a \task task, namely generating the gender-aware rewritten text, which is i) semantically relevant, ii) grammatically fluent, and iii) gender neutral.

\subsection{System Overview}
Figure~\ref{fig:system_archietecure} depicts the overall architecture of \system, which consists of a rewriting module, an administrative module, and multiple machine assistance services. The rewriting module offers annotators a user friendly interface for editing a given sentence with instant feedback. The feedback and revision histories in the interface are provided by the machine assistance services. Moreover, the administrative module provides administrators an interface for user management and assigning target \textit{tasks}, which are basically a set of sentences for rewriting, as \textit{jobs} to individual annotators.

\system is based on an easy-to-extend web-based framework that follows the Model-View-Controller~\cite{krasner1988description} software design pattern. The models are the wrappers of the databases (DB). The controller decides what should be displayed on the interfaces, which are considered as the views.
This flexible design enables various feedback providers to be easily plugged in and out, making it possible to support different text generation tasks. The front-end is developed with React\footnote{\url{https://reactjs.org}} that enables cross-platform support for major operating systems. 



\begin{figure*}[t]
    \centering
	\includegraphics[width=0.9\linewidth]{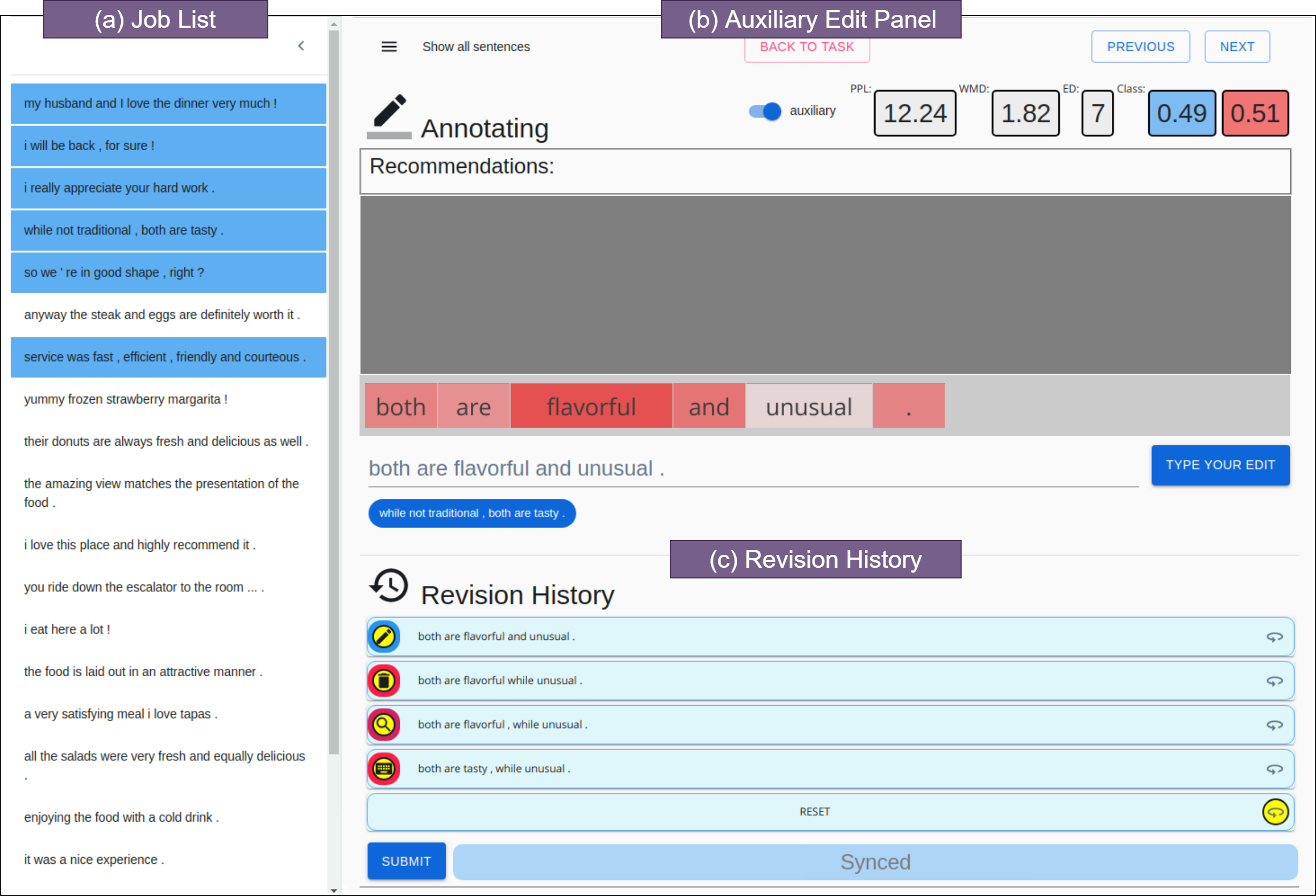}
	\vspace{-1mm}
	\caption{The Auxiliary Text Rewriting Interface is composed of (a) a job list, (b) an auxiliary edit panel and (c) a list of the revision history of current job.}
	\vspace{-2mm}
	\label{fig:rewriting_interface}
\end{figure*}

\subsection{Rewriting Interface}
Figure~\ref{fig:rewriting_interface} illustrates a screenshot of the annotator interface. In the left column, there is a list of jobs, which are the sentences assigned to the annotator. The completed jobs are marked in blue. An annotator starts with selecting an incomplete job from the job list, which will be shown in the auxiliary edit panel in the right column. We support two edit modes:
\begin{itemize}
    \item \textbf{Direct typing mode}: Annotators can directly type a whole sentence into the text input field. This mode is provided for the annotators who prefer typing to clicking. To save time, the original sentence is copied to the input field as default value.
    \item \textbf{Auxiliary mode}: Annotators can click on a word shown above the text input field, and choose one of the edit operations from a set,
    $S=\{$\textit{Word Typing}, \textit{Deletion}, \textit{Substitution}, \textit{Reordering}$\}$. If the annotator chooses \textit{Substitution}, he can select to show a list of words in the gray panel recommended by either word similarity or a pre-trained language model. In this mode, the annotator receives feedback from the upper right corner. Each feedback is a numerical score computed by a feedback provider based on the current sentence. After each edition, a record is added to the revision history below, with the corresponding edit operation and the modified sentences. The annotators are also allowed to roll back the sentences to a previous status by clicking the corresponding record in a history.
\end{itemize}


\subsection{Machine Assistance Services}
The machine assistance services in our system include feedback providers and word recommendation services. The machine assistance services can be categorized as sentence-level and word-level. 

At the sentence-level, we provide automatic sentence evaluation scores as feedback. In our current system, we consider evaluation metrics widely used in style transfer and obfuscation of demographic attributes~\cite{NAACL19_eval_style,zhao2018adversarially,fu2018style}. 
\begin{itemize}
    \item \textbf{PPL}. PPL denotes the perplexity score of the edited sentences based on the language model BERT\footnote{\url{https://github.com/google-research/bert}} \cite{NAACL19_bert}. 
    \item \textbf{WMD}. WMD is the word mover distance~\cite{ICML15_wmd} between the original sentence and the edited sentence based on Google's pre-trained Word2Vec model\footnote{\url{https://code.google.com/archive/p/word2vec}}. 
    \item \textbf{ED}. ED denotes the word edit distance between the original sentence and the rewritten sentence.
    \item \textbf{Class}. Class denotes the probability of the attribute value given the edited sentence. It is used to measure style transfer intensity or the degree of obfuscation. In our user study, we employ a transformer-based \cite{NIPS17_transformer} binary classifier trained on the \textbf{Gender} \cite{reddy2016obfuscating} corpus, which contains 2.6M balanced training samples.
\end{itemize}

At the word-level, we provide two word recommendation services for word substitution, which are based on word embedding similarity and language model, respectively. We include also a word-level feedback provider, which characterizes the contributions of individual words to the sentence-level classification results.
\begin{itemize}
    \item \textbf{Word Similarity Recommendation}. Given a selected word, this service recommends a list of words ranked by the cosine similarity computed based on pre-trained Google word embeddings.
    \item \textbf{Language Model Recommendation}. The services apply a pre-trained language model BERT to the context around the selected word to predict top-$k$ most likely words. 
    \item \textbf{Salience}. This module utilizes the sentence classifier trained on the Gender corpus to compute a salience score for each word. A salience score is defined as $S(X, i) = P(Y|X) - P(Y| X \setminus x_i)$, where $P(Y|X)$ denotes the probability of an attribute value $Y$ given the input sentence $X$, and $X \setminus x_i$ denotes the sentence $X$ excluding the $i$th word.
\end{itemize}


\section{User Study}
\label{sec:user_study}
We conduct empirical studies to demonstrate i) annotators fail to capture certain textual patterns leading to worse estimation accuracy than the classifier; ii) \system improves user engagement; iii) machine assistance consistently collects more references per sentence than asking annotators directly typing edited sentences. Both studies are based on the \textit{Gender}~\cite{reddy2016obfuscating} dataset, which consists of reviews from Yelp annotated with the gender of the authors. In the first study, we ask annotators to estimate the gender of authors given a sentence. In the second study, We consider a privacy-aware text rewriting task. We ask annotators to rewrite sentences that i) leak less gender information, ii) maximally preserve content; iii) are grammatically fluent.     


\subsection{Awareness of Gender Information}
\label{sec:human_vs_machine}
In the first study, we compare the accuracy of predicting gender information between two human annotators and the classifier\footnote{We use a linear SVM model trained on \textbf{Gender}.}. Both of them predict the authors' gender of 300 sentences randomly sampled from the test set. Human annotators obtain merely 66.0 of accuracy on average, while the classifier achieves 77.3. 
We have carefully investigated the prediction results and the sampled sentences. We found out that it is indeed difficult for humans to estimate correctly the authors' gender based on a short piece of text, e.g.,``the food is delicious'' and ``the people were nice''. Both examples are perceived as neutral for our annotators.
Apart from human failure to capture weak associations between certain textual patterns and gender, we conjecture that the bias in the corpus may help the classifier achieve better performance.

\subsection{User Engagement}
\label{sec:user_engagement}
In this study, three graduate students are invited to rewrite 100 sentences randomly selected from the test set of the Gender corpus. All students take two steps to rewrite each sentence:
\begin{enumerate}
    \item In the \textit{direct typing mode}, type the edited sentence directly in the input field .
    \item In the \textit{auxiliary mode}, improve the edited sentence from the first step when necessary. The annotators are instructed that i) it is fine to leave the sentences as they are if feedback do not provide useful clues; ii) all feedback and recommendations are machine generated, thus not perfect.
\end{enumerate}
We consider the two-step approach to compare the differences between the two modes while minimizing individual differences between annotators.

\begin{figure}[t]
    \centering
	\includegraphics[width=0.90\linewidth]{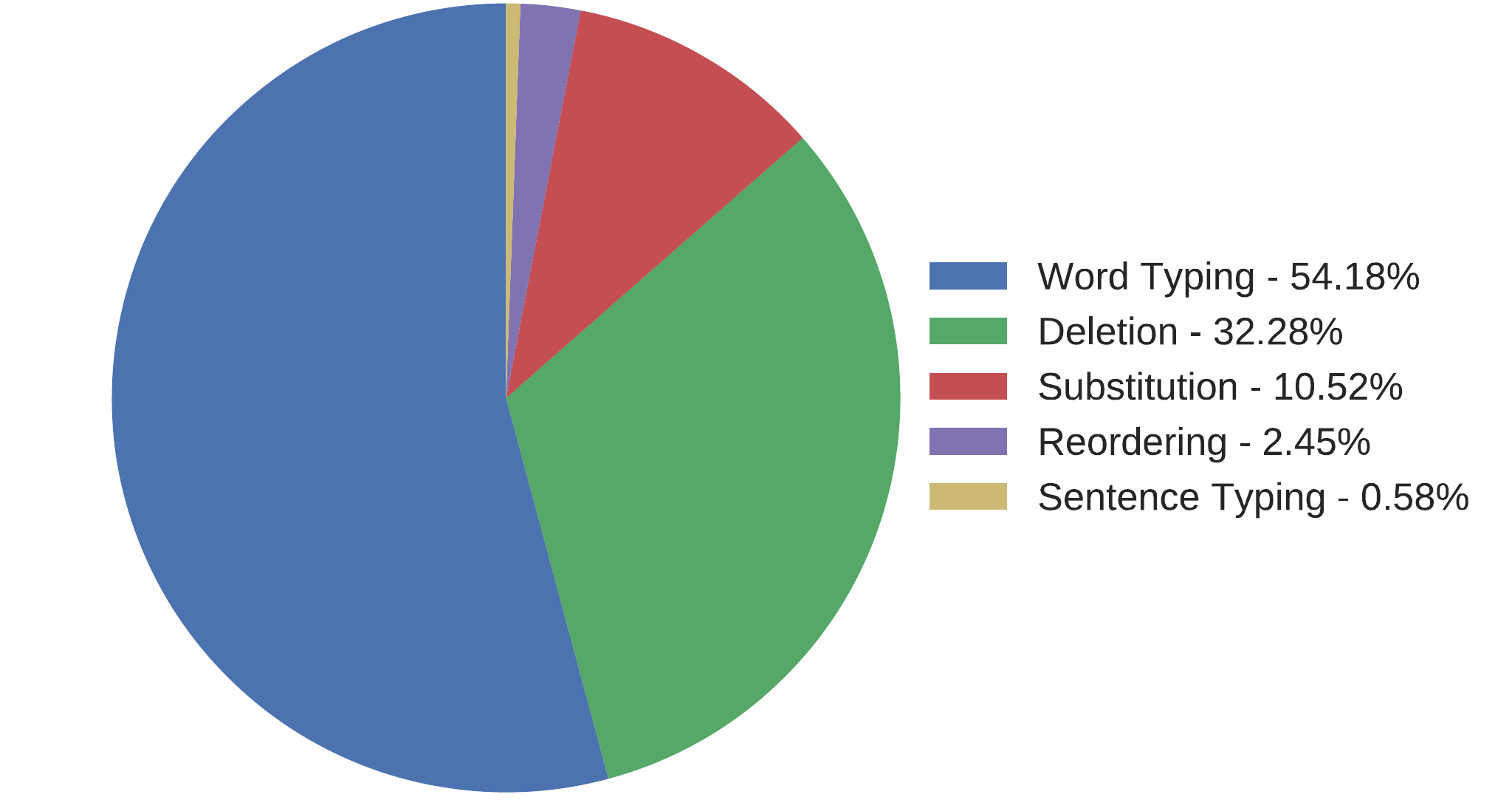}
	\vspace{-2mm}
	\caption{Distribution of operations in revision history by Word Typing, Deletion, Substitution, Reordering and Sentence Typing.}
	\label{fig:opt_dist}
\end{figure}

We analyze the revision history collected in the second step, and found out that feedback indeed leads to significant improvement of user engagement. In the second step, 89.67\% of the sentences were modified in the auxiliary mode. The average number of edit operations in the second step is 4.63, showing the willingness of writers to further edit the text under auxiliary mode. The distribution of edit operations is illustrated in Figure~\ref{fig:opt_dist}, \textit{word typing} and \textit{deletion} are clearly the most popular edit operations. Word recommendation services are also effective, contributing more than 10\% of the new edits in the auxiliary mode.

The references collected in the second step result in less leakage of gender information than the ones in the first step. We measure the leakage of gender information by applying the transformer-based classifier on references collected in both steps. We compute averaged \textit{entropy} score, $-\sum_i p_i \log p_i$, based on the predication of each class $p_i$.
Higher \textit{entropy} indicates better obfuscation of gender. The sentences collected in the first step and the second step achieve 0.347 and 0.535 respectively. The entropy of the sentences collected in the first step is just 0.027 better than that of the original sentences. 

We further investigate the revision histories, and find more gold references per sentence in the second step than in the first step. We consider semantically relevant and grammatically fluent sentences as valid references. The average number of the valid references generated in auxiliary mode is 3.79, while we can merely obtain one reference per sentence in the direct typing mode. 



\section{Conclusion and Future Work}
In this paper, we demonstrate our auxiliary text rewriting tool \system{} to collect gold references for \task, assisted with word-level revision histories and task-specific instant feedback. In the future, 
we will apply \tool to collect high-quality benchmarks for \task. 

\section*{Acknowledgement}

This project is supported by the partnership between ANU and Data61/CSIRO. We also gratefully acknowledge the funding from Data61 scholarship that supports Qiongkai Xu and Chenchen Xu's PhD research.


\bibliography{emnlp-ijcnlp-2019}
\bibliographystyle{acl_natbib}

\newpage



\end{document}